\documentclass[10pt,twocolumn,letterpaper]{article}

\usepackage{cvpr}
\usepackage{times}
\usepackage{epsfig}
\usepackage{graphicx}
\usepackage{amsmath}
\usepackage{amssymb}
\usepackage{multirow}
\usepackage{adjustbox}
\usepackage{tabularx}
\usepackage{subcaption}

\usepackage[breaklinks=true,bookmarks=false]{hyperref}

\cvprfinalcopy 

\def\cvprPaperID{****} 

\begin{document}

\title{Solar Power Plant Detection on Multi-Spectral Satellite Imagery using Weakly-Supervised CNN with Feedback Features and m-PCNN Fusion}

\author{Nevrez Imamoglu  \thanks{ These authors equally contributed to this work. \newline {\addtolength{\leftskip}{5mm} \indent This paper is based on the results obtained from a project commissioned by the New Energy and Industrial Technology Development Organization (NEDO), Japan. } }  \and Motoki Kimura \footnotemark[1] \and Hiroki Miyamoto\and Aito Fujita\and Ryosuke Nakamura\\
National Institute of Advanced Industrial Science and Technology\\
AIST Tokyo WaterFront, Tokyo, Japan\\
{\tt\small \{nevrez.imamoglu, kimura.motoki, miyamoto-hrk-tomeken, fujita.713, r.nakamura\}@aist.go.jp}
}

\maketitle

\begin{abstract}
Most of the traditional convolutional neural networks (CNNs) implements bottom-up approach (feed-forward) for image classifications. However, many scientific studies demonstrate that visual perception in primates rely on both bottom-up and top-down connections. Therefore, in this work, we propose a CNN network with feedback structure for Solar power plant detection on middle-resolution satellite images. To express the strength of the top-down connections, we introduce feedback CNN network (FB-Net) to a baseline CNN model used for solar power plant classification on multi-spectral satellite data. Moreover, we introduce a method to improve class activation mapping (CAM) to our FB-Net, which takes advantage of multi-channel pulse coupled neural network (m-PCNN) for weakly-supervised localization of the solar power plants from the features of proposed FB-Net. For the proposed FB-Net CAM with m-PCNN, experimental results demonstrated promising results on both solar-power plant image classification and detection task.
\end{abstract}

\section{Introduction}

With the gradual increase in global warming of our planet Earth, investments on clean energy sources such as solar power plants have been increasing all over the world. For example, the photovoltaic capacity had expected to exceed 270 gigawatts worldwide between 2010 and 2016 as expressed in [1, 2]. These solar power plants are visible through satellite imagery for allowing visual mapping and labelling of the plants. Since satellites have been collecting data over years, it may also possible to observe temporal maps or labels regarding the change in clean energy development in solar power plants.

\begin{figure}[t]
\begin{center}
\fbox{ \includegraphics[width=0.95\linewidth]{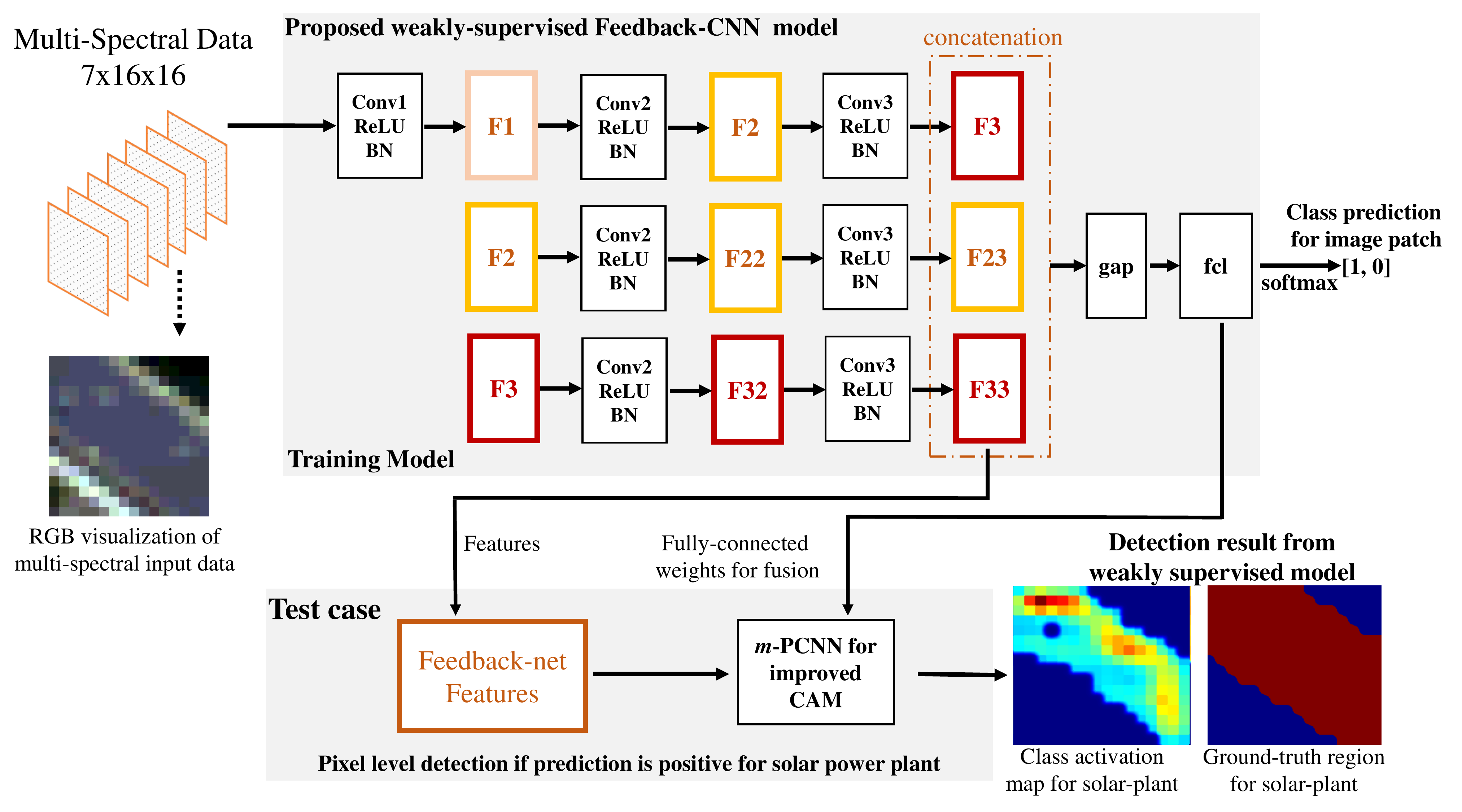}}
\end{center}
   \caption{\textbf{Training model:} Weakly supervised CNN Feedback-Net (FB-Net) model for solar power plant classification on multi-spectral imagery. The model includes three convolution layers (each convolution layer followed by ReLU and batch normalization) with feedback from features F2 and F3 to second layer, global average pooling layer (gap), and fully connected layer (fcl)  \textbf{Test case:} Trained FB-Net features F3, F23, and F33 are used for class activation mapping (CAM). Instead of CAM in [17] as weighted linear combination, multi-channel Pulse Coupled Neural Network (m-PCNN) is used for feature fusion with fcl weights to obtain CAM.}
\label{fig:fig1}
\end{figure}

Recent advancements of deep learning [3-5] enabled researchers to develop many models, especially convolutional neural networks (CNNs), for classification or detection tasks on satellite imagery [2, 6-9]. However, to the best of our knowledge, simple yet successful implementation of CNN model proposed by Ishii et al. [2] is the only base line work on Solar Power Plant classification on a large-scale (covering Japan) multi-spectral satellite data. Using Landsat-8 data for Japan, the network [2] classifies small patches (16$\times$16 spatial resolution) and labels these patches with likelihood of being solar power plant. And, the system is running online on LandBrowser interface [10]. However, in their work [2], detection by classification refers to labelling all pixels in the whole patch as the class prediction obtained from classification of the input data. So, we investigate pixel classification by using their model as a baseline to extend and improve solar power plant detection task. Certainly, current state-of-the-art models for recognition of remote sensing data such as Alex-Net [3] based CNN model OverFeat [9] can also be used by resizing 16$\times$16 image patches to compatible input sizes for these models (e.g. 231$\times$231 or 227$\times$227). However, considering the model complexity of [3] and [9], this  is not flexible and computationally more expensive for processing one Landsat-8 scene with approximately 7500$\times$7500 pixels. Also, as demonstrated in [2] and in our experimental results section, model in [9] applied to solar power plant classification does not perform better than relatively simple model of Ishii et al. [2] CNN model (i.e. three convolution layers followed by a fully connected layer).

On the other hand, conventional CNN implementations generally use feed-forward direct connections for layer structures to do classification and detection tasks [3-9]. However, as stated in [11], many scientific studies [12, 13] on visual perception in primates have shown important signs of both bottom up and top-down connections in visual cortex areas. Inspired by these researches, CNN models with feedback unit or recurrent systems have been tried in some vision applications [11, 14-16] with the recent deep learning tools. These works [11, 14-16] use top-down information as modulator signal on the CNN layers or use them as a part of selective attention process to adjust neural activations between layers. However, despite being less plausible from the biological perspective, another way of implementing feedback on CNN is to use higher contextual features as input to lower CNN layers, and then, apply feed-forward direct computation. With these feedback structure, it is possible to obtain richer feature representation (e.g. more class dependent feature activations) before the decision layer by providing relatively low-level and high level contextual features. Therefore, in this work, we describe a CNN model based on feedback features to achieve solar power plant classification and detection on multi-spectral imagery. 

In addition, as the dataset shared by [2] does not include pixel-wise annotation for training data, it is not possible to try fully-supervised models such as FCN [5] or similar end-to-end pixel-wise training models. Therefore, we investigate the use of localization methods on weakly-supervised CNNs through taking advantage of class activations map approaches such as CAM [17]. However, we consider that class dependent activation of a pixel should consider the activations around the surrounding pixels while fusing feature channels with respect to weights learned or obtained from the CNN model. Therefore, to improve the detection performance of class activation maps, we propose the use of pulse-coupled neural networks (PCNN) for feature fusion by integrating it as an improved CAM approach. PCNN feature fusion not only considers weighted average of channels but also it considers activations of each pixel while taking surround pixel activations to consideration. 

\textbf{With the stated purpose, our contributions in this paper are the following:}
\begin{quote}

i) We propose a feedback network (Fig.1) for Solar power plant classification on middle-resolution (30 meters/pixel) satellite images. To express the strength of the top-down connections, we introduce feedback structure (FB-Net) to Ishii et.al. [2] baseline CNN model (I-Net) used for solar power plant classification. 

ii) As given in Fig.1, we use global average pooling for features before fully connected layer as in [17] to obtain weights in feature fusion for weakly-supervised class  activation mapping [17,18]. However, in test phase, we propose use of multi-channel PCNN (m-PCNN) [19-21] based feature fusion to improve class activation mapping (CAM) [17, 18] to our FB-Net as a weakly-supervised localization of class depended features for solar power plants. m-PCNN provides improved feature fusion due to spatial connections of the pixels, which effects final pixel class activations.

\end{quote}

For image patch classification performance evaluation, we use Intersection over Union (IoU) metric as in [2]. For detection performance of the proposed m-PCNN based CAM model with our FB-Net, we make quantitative validation by doing analysis and comparison on variations of I-Net [2] and our proposed FB-Net, which are demonstrated in experimental results. To do so, we apply simple feature averaging, CAM [17], and Gradient based CAM [18] on CNN features for both I-Net [2] and our FB-Net. As detection evaluation metric for class activation maps, we use Area Under Curve (AUC) computed from Receiver Operating Characteristic (ROC) curve. ROC curve is obtained through applying various thresholds to the activation maps. For the proposed FB-Net CAM with m-PCNN (Fig.1), experimental results demonstrated promising results on both mega-solar classification and detection task based on the IoU and AUC metrics respectively.

The outline of the paper follows: Section 2 is the brief explanation for the dataset used in this work, Section 3 describes the proposed weakly-supervised FB-Net-CAM with m-PCNN integration for classification and detection, Section 4 gives some background information on related work for comparison in experiments,  and Section 5 demonstrates the experimental results, finally, some concluding remarks are given.

\section{Dataset}

To train and test the models, we use the dataset which was introduced and provided by Ishii et al. [2]. For this dataset, publicly available multi-spectral satellite images captured by Landsat 8 and power plant database are used. In the following sub-section, we explain the overview of dataset from Landsat 8 imagery.

\subsection{Landsat 8 imagery}

Landsat 8 is an American Earth observation satellite which has been operated since 2013 [2, 22, 23]. It observes whole surface of the Earth with 11 bands of different wavelength and spatial resolutions [22, 2]. We show an overview of imaging sensors and bands of Landsat 8 in Table 1 [22, 23, 2]. 

As explained in [2], we also used the first 7 bands having same spatial resolution and mostly no overlap in wavelength. The entire dataset was constructed from 20 satellite images of Japan captured in 2015, and each of the images captures roughly an area of 170km$\times$183km [2].

\begin{table}[h]
\begin{center}
\begin{tabular}{l|c|c|c}
\hline
Sensor & Band & Wavelength ($\mu$m) & Resolution (m)\\
\hline\hline

\multirow{9}{*}{OLI} & 1 & 0.43-0.45 &30\\
& 2 (B) & 0.45-0.51 &30 \\
& 3 (G) & 0.53-0.59 &30 \\
& 4 (R) & 0.64-0.67 &30 \\
& 5 & 0.85-0.88 &30 \\
& 6 & 1.57-1.65 &30 \\
& 7 & 2.11-2.29 &30 \\
& 8 & 0.53-0.68 &15 \\
& 9 & 1.36-1.38 &30 \\
\hline
\multirow{2}{*}{TIRS} & 10 & 10.60-11.19 &100\\
& 11 & 11.50-12.51 &100\\
\hline
\end{tabular}
\end{center}
\caption{Observation wavelength and spatial resolution of Landsat 8 imaging sensors. OLI corresponds to visible and near-infrared light, while TIRS corresponds to thermal infrared light}
\end{table}

\subsection{Solar power plant data}

In [2], Solar power plants are manually annotated with a polygon in the satellite images. Then, small patches with 16$\times$16 spatial resolution are extracted. A sample patch is labelled as mega-solar if it covers more than 20\% solar panel area obtained by the manually drawn polygons [2]. Then, to use these satellite imageries and manually arranged ground truth images as input data to CNNs, each imagery was divided into cells with the size of 16$\times$16 pixels. Each cell covering an area of 480m$\times$480m is treated as the input data for CNN model.

Because the Landsat-8 [22, 23] satellite image is too coarse for solar panels or relatively small scale solar power plants, such solar panels are not feasible for manual labelling for generating training and test data. Therefore, the authors in [2] create positive samples if only the power plants whose output is greater than 5MW [2]. With these conditions, we obtained the training and test data from [2] with sample numbers are given in Table 2.

\begin{table}[h]
\begin{center}
\begin{tabular}{r|r|r}
\hline
\multicolumn{2}{ c|}{Dataset type} & \#patches\\
\hline\hline
\multirow{2}{*}{training} & $\mathbf{P}_{train}$ & 4851\\
& $\mathbf{N}_{train}$ & 320000\\
\hline
\multirow{2}{*}{test} & $\mathbf{P}_{test}$ & 105\\
& $\mathbf{N}_{test}$ & 802666\\
\hline
\end{tabular}
\end{center}
\caption{Details of the train and test partitions of the dataset. Each patch has 7 channels with the size of 16$\times$16 pixels. The number of positive samples P and negative samples N for each part train and test are given}
\end{table}

\section{Proposed Method}

\subsection{Feedback CNN Model for Solar Power Plant Classification}

In this paper, we propose a model which has feedback paths by using I-Net (Ishii et al. [2]) as the baseline network. We refer to our model as FB-Net in this paper. FB-Net has 3 convolutional layers followed by a fully connected layer like I-Net but it has 2 feedback paths inside the model as shown in Fig.1. The features output from the second convolutional layer (F2 features in Fig.1) goes back to the second convolutional layer and then goes through the third convolutional layer. This feedback path is illustrated as the middle path of the trained model in Fig.1. Likewise, the features output from the third convolutional layer (F3 features in Fig.1) goes through the second convolutional layer and then the third convolutional layer again. This feedback path is the bottom path of the trained model illustrated in Fig.1. The features output from the top, middle, and bottom paths (see Fig.1, F3, F23, and F33 features respectively) are all concatenated before the fully connected layer. Note that all the second convolutional layers shown in Fig.1 share the same weights as well as the third convolutional layers. These features are processed by global average pooling followed by a fully connected layer that output the prediction result for classification task with soft-max.

\subsection{Solar Power Plant Detection with Weakly Supervised Feedback-Net with m-PCNN based Class Activation Mapping}

I-Net in [2] uses pixel-level labelling of solar power plants results from classification. So, if an input data with 16$\times$16 spatial resolution is predicted as solar power plant, detection by classification in [2] assigns labels to all pixels covering 480m$\times$480m as part of solar power plants. However, this very coarse labelling cannot result in reliable accuracy because one pixel is around 30 meters on Landsat-8 multi-spectral data even though it may show reasonable and acceptable solar-power plant areas on a data with very large spatial resolution. In addition, approaches such as CAM and Grad-CAM does not consider the activations of surrounding regions which may results in false positives. Therefore, by taking advantage of pulse-coupled neural network (PCNN) based class activation mapping, we introduce a solar power plant detection framework with pixel-level segmentation ability.

PCNN is a promising algorithm on information fusion. As an unsupervised neural network model, PCNN demonstrated its usability in various image processing applications such as information fusion, segmentation, etc. [19-21]. The common PCNN model is based on Eckhorn model which introduces the cat visual cortex [24]. This model is modified for PCNN to use in digital applications [24, 19-21]. In this paper, we take advantage of multi-channel PCNN (m-PCNN) model proposed by [19], and we apply m-PCNN for feature fusion to create class activation map of solar power plants. Iterative process of m-PCNN (see Fig.2) can be formulated through Eq.1 to Eq.4 as [19, 20]:

\begin{equation}
\label{eq:Eq1}
\begin{aligned}
	\resizebox{.9\hsize}{!}{$\mathit{H}_{i,j}^{k}[n]=  e^{-\alpha_{H}}\mathit{H}_{i,j}^{k}[n-1] + V_{H}\left(\mathbf{W}*\mathit{Y}_{i,j}[n-1]\right) + \mathit{S}_{i,j}^{k}$} \\
	\resizebox{.75\hsize}{!}{$\mathit{W}_{i,j}= \frac{1}{\left( \left( i-c_{x} \right)^{2} + \left( j-c_{x}  \right)^{2}  \right)^\frac{1}{2}} \text{, and} \mathit{W}_{c_{x},c_{y}}= 0$}
\end{aligned}
\end{equation}
\begin{equation}
\label{eq:Eq2}
	\mathit{U}_{i,j}[n]=\Pi_{k=1}^{K} \left( 1+\beta^{k}\mathit{H}_{i,j}^{k}[n] \right)
\end{equation}
\begin{equation}
\label{eq:Eq3}
	\mathit{Y}_{i,j}[n]=\begin{cases}
			    1, &  \mathit{U}_{i,j}[n]>\mathit{T}_{i,j}[n-1]\\
			    0, & \text{otherwise}
  \end{cases}
\end{equation}
\begin{equation}
\label{eq:Eq4}
	\mathit{T}_{i,j}[n]= e^{-\alpha_{T}}\mathit{T}_{i,j}[n-1]+ V_{T}\mathit{Y}_{i,j}[n]
\end{equation}
\noindent where $k$ = \{1,...,96\} refers to the input channels (CNN features extracted from FB-Net; i.e. F3, F23, and F33 in Fig.1 are concatenated), $\mathbf{H}$ is the external stimulus from the feed function in (Eq.1) with input stimulus $\mathbf{S}$, $\beta^{k}$ is the weight showing the importance of the $k^{th}$ data channel, $\mathbf{W}$ is the constant synaptic linking weights, "*" is the convolution operation, $\mathbf{U}$ is the combination of feeding and linking process, $\mathbf{U}$ also expresses the internal state of the neuron at iteration n, $\mathbf{Y}$ is the fired neurons that are defined by the dynamic threshold $\mathbf{T}$, $V_{H}$ and $V_{T}$ are used as scaling parameters, and $\alpha^{H}$ and $\alpha^{H}$ are the time constants. When the all neurons (pixels) are processed through the fusion of Feedback-Net CNN features, the iteration process from Eq.1 to 4 terminates. And, $\mathbf{U}$ at the last iteration is the fusion result as our output for mega-solar detection. $\mathbf{H}$ and $\mathbf{S}$ is initialized as the feature channels.

\begin{figure}[t]
\begin{center}
\fbox{ \includegraphics[width=0.95\linewidth]{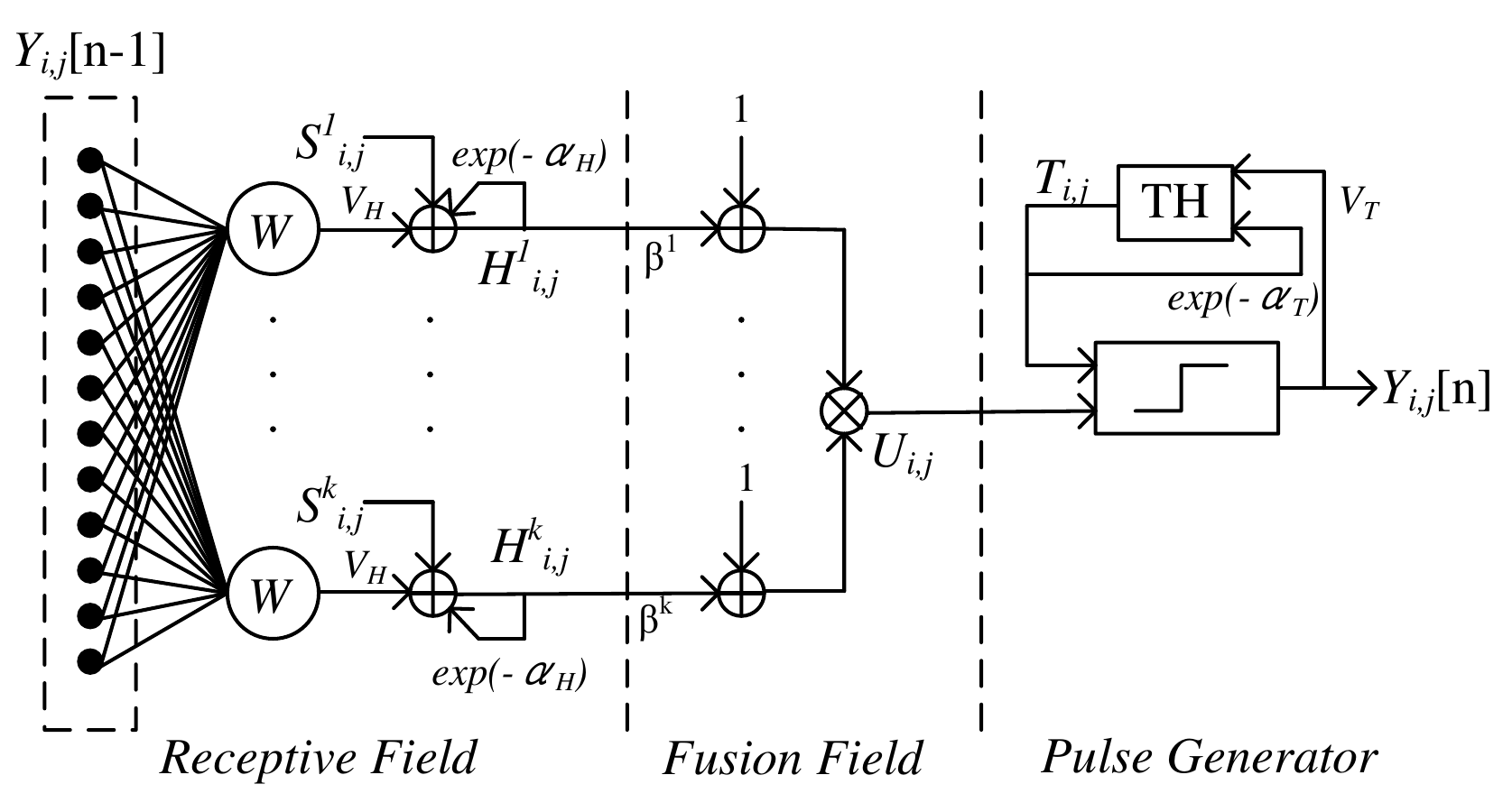}}
\end{center}
   \caption{Structure and work flow of an m-PCNN neuron}
\label{fig:fig2}
\end{figure}

To use m-PCNN to obtain class activation map on CNN features, we adaptively assign weighting factor $\beta_k$ based on the class dependent weights obtained from the fully connected layer (see Fig.1) connected to global average pooling [17]. Proposed FB-Net CAM with mPCNN results in very promising detection results. Section 4 explains some background related works, and  Section 5 shows the performance analysis of the proposed classification and detection tasks on test data.

\section{Related Works}

In Ishii et. al's [2] model (I-Net), they propose a detection by classification approach on 16x16 spatial resolution patches. In their work [2], detection by classification refers to labelling all pixels in the whole patch as the class prediction obtained from classification of the input data. Therefore, we use I-Net as our base line model for comparison with variations of it for classification and detection.

I-Net has 3 convolutional layers followed by a fully connected layer as shown in Fig.3 Parameters and transfer function of each layer are shown in Table 3 ReLU is used as transfer function after all convolutional layers. All of the convolutional layers have 3$\times$3 kernels and no padding. Unlike most of the standard models, no pooling layer is used and all convolutional layers have a stride of 1 due to the small size of input images. Despite the architecture of the model is relatively simple for purpose of near real time processing of large satellite images, it shows reasonably high accuracy for the classification. The classification accuracy of I-Net is around 0.52 in metric of intersection over union after tuning the threshold  with validation [2].

\begin{figure}[h]
\begin{center}
\fbox{ \includegraphics[width=0.95\linewidth]{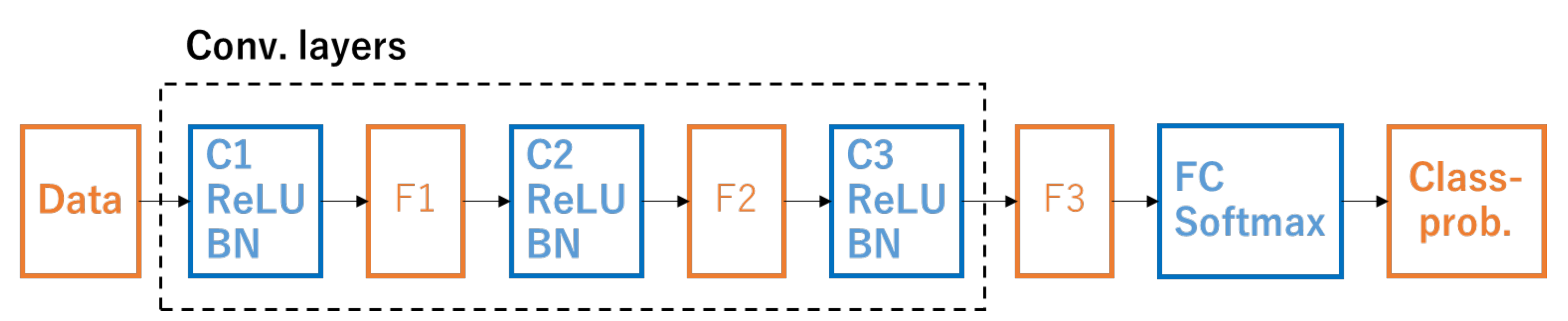}}
\end{center}
   \caption{Architecture of I-Net model. “C”: convolutional layer, “FC”: fully connected layer, “F”: features, “BN”: batch normalization}
\label{fig:fig3}
\end{figure}

\begin{table}[h]
\begin{center}
\begin{adjustbox}{width=0.475\textwidth}
\begin{tabular}{lccc}
\hline
layer & kernel size & input size & transfer function\\
\hline\hline
input & - & 7x16x16 & - \\
convolution & 3x3 & 32x14x14 & ReLU \\
convolution& 3x3 & 32x12x12 & ReLU \\
convolution& 3x3 & 32x10x10 & ReLU \\
fully connected& - & 2 & softmax \\
\hline
\end{tabular}
\end{adjustbox}
\end{center}
\caption{Detailed architecture of I-Net. All convolutional layers have kernel with the size of 3$\times$3 and outputs the feature maps with the channel of 32}
\end{table}

\subsection{Related works for Detection}
Since baseline model I-Net is not using pixel level detection, we apply several approaches on I-Net and proposed Feedback-Net from literature for detection of mega-solar power plants. This allows us to make extensive experimental evaluation and comparison on I-Net and FB-Net trained models for classification and detection tasks. In this section, we introduce state-of-the-art methods applied on I-Net for obtaining region proposals on detection task such as CAM [17], Grad-CAM [18], and simple feature averaging. Then, we evaluate each approach and show that the proposed detection method surpasses I-Net and its variations for pixel level labeling of the solar power plants. We also try these approaches on FB-Net for comparison.

\paragraph{Class Activation Mapping:}
One of the related approaches for detection is the Class Activation Mapping (CAM) [17] used for region detection or pixel level object localization tasks. This approach modifies image classification CNN architectures by replacing connection between fully-connected layers and convolutional layers by global average pooling layer (Fig.4), to achieve class-specific feature maps [17].

To apply CAM to I-Net as one of the detection model example for comparison, we add a global average pooling layer just before the final fully connected layer of I-Net as shown in Fig. 4 and train the model from scratch. 

\paragraph{Gradient-weighted Class Activation Mapping:}

Gradient-weighted Class Activation Mapping (Grad-CAM) [18] is a strict generalization of CAM, which uses the class-specific gradient information flowing into the final convolutional layer of the model to produce a course localization map of the important regions in the input image for each class. Unlike CAM [17], Grad-CAM [18] does not require any model modifications.

\begin{figure}[t]
\begin{center}
\fbox{ \includegraphics[width=0.95\linewidth]{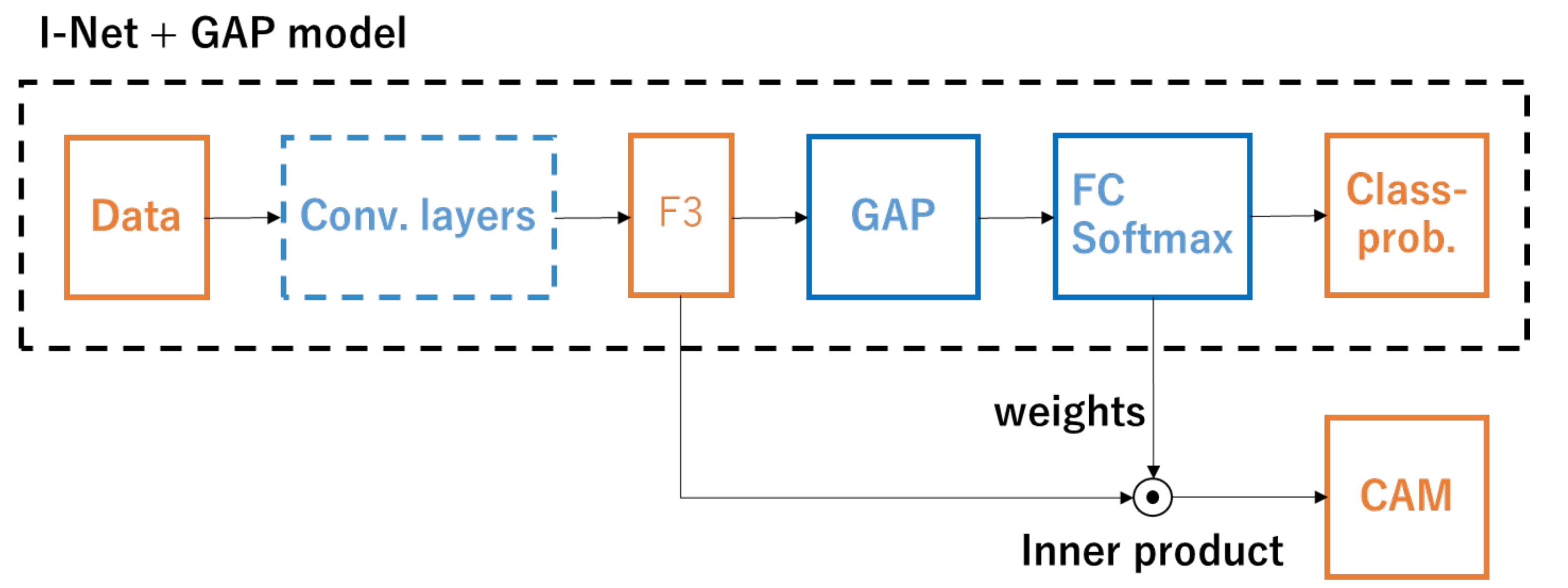}}
\end{center}
   \caption{Modified I-Net architecture for Class Activation Mapping. A global average pooling (GAP) layer is added before the fully connected layer. CAM can be calculated as the inner product of feature maps output by convolutional layers and weights of the fully connected layer.}
\label{fig:fig4}
\end{figure}

\begin{figure}[t]
\begin{center}
\fbox{ \includegraphics[width=0.95\linewidth]{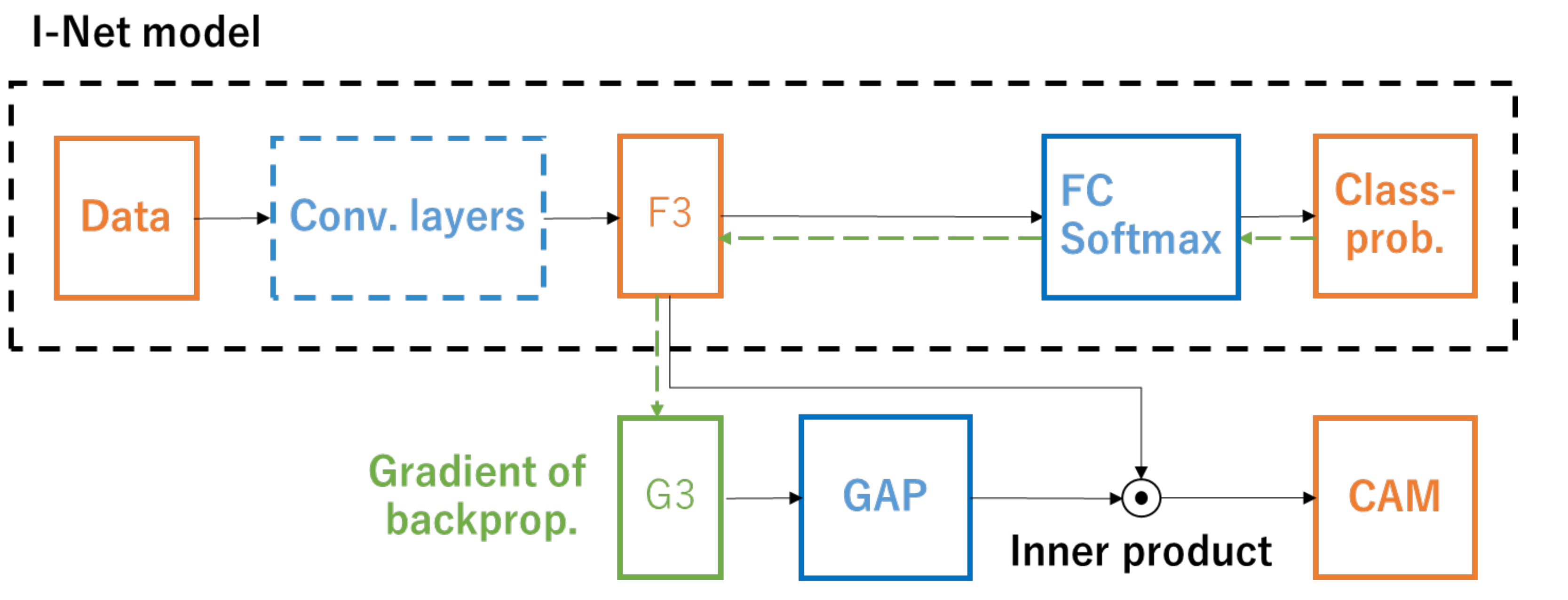}}
\end{center}
   \caption{Grad-CAM with I-Net model. Grad-CAM can be calculated as the inner product of F3 and weights from G3 by applying GAP on G3.}
\label{fig:fig5}
\end{figure}

When applying Grad-CAM, we use output features and gradients of the last convolutional layer, i.e. the convolutional layer just before the final fully connected layer. We show Grad-CAM with I-Net in Fig. 5 as one of the detection model example for comparison. Grad-CAM is calculated as the inner product of features output by the last convolutional layer and gradient of the features after applying global average pooling. This gradient work as the weights of the fully connected layer of CAM model.

\paragraph{Feature averaging:}

We also try feature averaging as a simple class activation mapping way for detecting solar power plants. To create region proposal for mega-solar, we simply calculated the average of output features of the last convolutional layer in channel-direction.

\section{Experiments}

\subsection{Training of the models}

\paragraph{Batch normalization and dropout layers:}
We train the models by using the dataset of 16$\times$16-pixel image patches with corresponding annotations which we already explained in Section 2 using cross-entropy loss. To train the models efficiently from scratch, we use Batch Normalization layers [25, 2] after ReLU function of each convolution layer. 

Batch Normalization layers are only used when training, and these layers become a fixed linear transformation in test and validation phase. As in [2], we also use a Dropout layer [26] that sets the output after the third convolution layer and before the fully connected layer to 0 with 50\% possibility, independently at each pixel of the output. This layer reduces overfitting of the models to the training data [2, 26]. We use SGD (Stochastic Gradient Descent) with fixed learning rate of 0.01 and momentum of 0.9 to optimize weights of the models.

\paragraph{Balanced mini-batch sampling:} As described in [2], despite augmenting the positive samples and subsampling the negative samples, training dataset still has a large bias between positives and negatives. When training model, we get mini-batches with the size of 128 samples from the training set. Each mini-batch consists of 8 positives and 120 negatives so that there are always positive and negative samples for every mini-batch. With this balanced sampling of negative and positive samples, we prevent the model training from being biased by negative samples despite the large number.

\subsection{Experimental Results for Classification}

In this section, we compare solar-power plant classification performance (IoU metric) of proposed Feedback-Net (see Fig.1) with I-Net [2]. We also added CNN model of Pennati et al. [9] and Support Vector Machine (SVM) based classification results demonstrated in [2]. In addition, we also tried I-Net with global average pooling (gap) for CAM and our FB-Net without gap layer to see if classification accuracy is decreasing on I-Net and FB-Net by adding GAP as referred in CAM paper [17]. After training each CNN model up to 4,000 epochs, we evaluate each model in both classification and detection accuracy with the best performing epoch based on intersection over union (IoU) for each model respectively using test data with 105 positive and 802666 negative samples of image patches. Regarding solar power plant classification performance, comparison of proposed FB-Net to SVM in [2], Pennati [9], I-Net [2], I-Net with gap, and our FB-Net without gap are given in Table 4.

\begin{table}[h]
\begin{center}
\begin{adjustbox}{width=0.475\textwidth}
\begin{tabular}{lcccc}
\hline
Model &  TP rate & TN rate & IoU\\
\hline\hline
SVM [2]	    & n/a            & n/a             & 0.23 \\
Penatti [9]     & n/a            & n/a             & 0.39 \\
I-Net             & 0.923810  & 0.999796  &  0.36 \\
I-Net+gap   & \textbf{0.933333}  & 0.999722  & 0.30 \\
FB-Net w/o gap           	& 0.923810  & 0.999779  & 0.34 \\
\textbf{FB-Net} & 0.838095  & \textbf{0.999935}  & \textbf{0.56} \\
\hline
\end{tabular}
\end{adjustbox}
\end{center}
\caption{Intersection over Union based performance comarison of proposed FB-Net with SVM [2], Penatti [9], I-Net [2], I-net with gap layer, and our FB-Net without gap layer}
\end{table}

It should be noted that our test accuracy on I-Net [2], I-Net+gap, FB-Net w/o gap, and proposed FB-Net (Fig.1) are results directly using training model without validation data, which can help to find optimal threshold on soft-max output value for decision. We can improve classification accuracy even more by using validation data to decide optimal decision threshold on the model predictions. However, in this work, we would like to focus more on the detection performances of these models for the true positive cases, which is explained in the following section. As demonstrated in [2], I-Net [2] yields 0.52 IoU accuracy when validation data is used, which is outperforming SVM and Pennati [9] model for solar-power plant classification. In our experiments, I-Net [2] accuracy decreased when gap layer is added to the network for obtaining CAM, which is also an observation of [17] proposing CAM approach. On the other hand, interestingly, our proposed FB-Net has better IoU performance than FB-Net without gap model. 

Since the training and test data is largely biased to negative samples, very small difference, or error on True Negative rate, may result in a significant change on the IoU result despite having high classification accuracy both on True Positive (TP) and True Negative (TN) rate. Therefore, IoU performance highly dependent on the accuracy of TN rate compared to TP rate. For example, proposed FB-Net demonstrated the best performance regarding IoU because it has the highest TN rate regardless of its lower TP rate accuracy compared to other approaches. However, FB-Net without gap layer has worse IoU performance than both original I-Net [2] without gap layer and proposed FB-Net using gap layer.

\newcolumntype{C}{>{\small\centering\arraybackslash}X}
\begin{figure*}
\begin{center}
	\fbox{\parbox{\dimexpr\linewidth-2\fboxsep-2\fboxrule\relax}
		{
		\begin{tabularx}{0.985\textwidth}{CCCCCCCCCC}
		         (a) & (b) & (c) & (d) & (e) & (f) & (g) & (h) & (i) & (j)
		\end{tabularx}
		\\
		\begin{tabular}{c}
			{\includegraphics[width=0.985\linewidth]{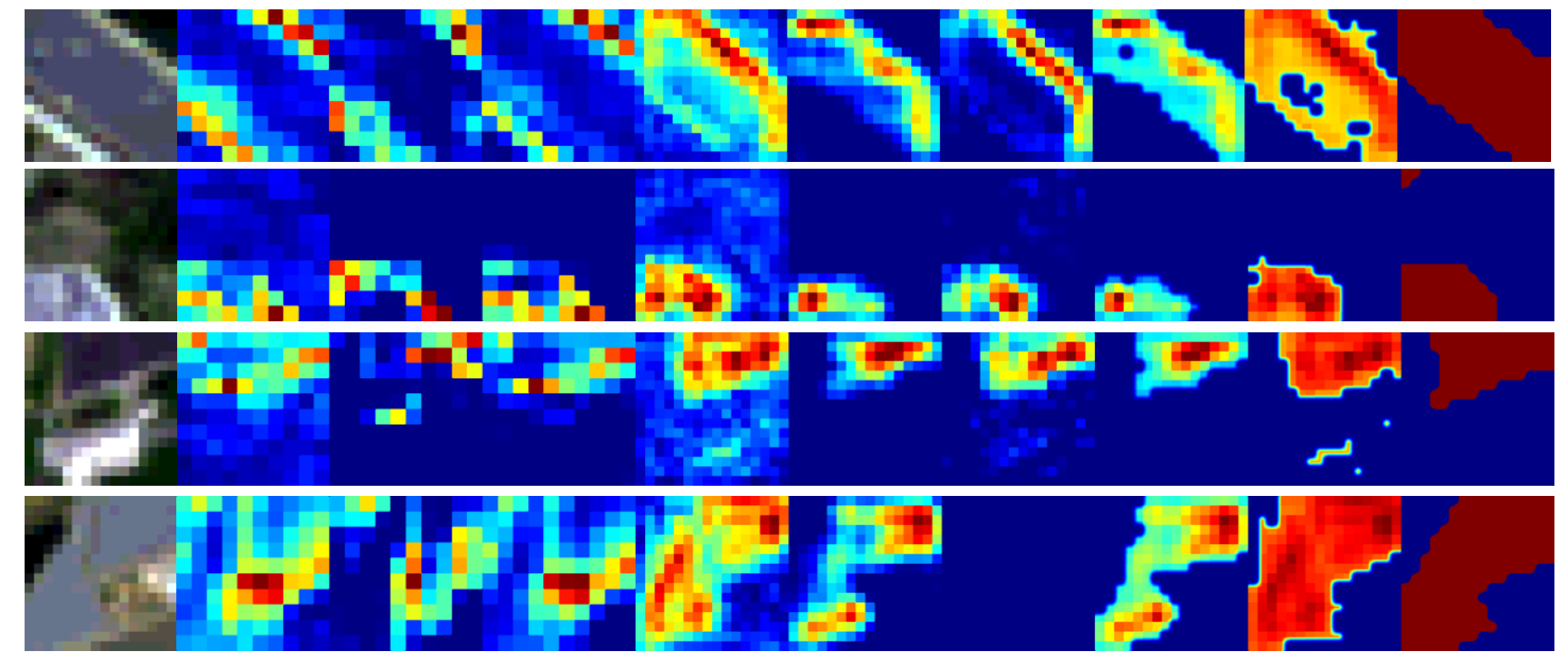}}
		\end{tabular}
		}
	}
\end{center}
   \caption{Solar power plant detection results of (a) sample images (RGB visualization of multi-spectral data) (b) I-Net [2] feature averaging, (c) I-Net+gap, (d) I-Net using Grad-CAM in [18], (e) feature averaging on FB-Net w/o gap, (f) FB-Net using CAM in [17], (g) FB-Net w/o gap using Grad-CAM [18], (h) Proposed FB-Net with mPCNN based CAM, (i) FB-Net w/o gap layer with mPCNN based Grad-CAM, (j) ground-truth}
\label{fig:fig6}
\end{figure*}

\subsection{Experimental Results for Detection}

We compared the detection performance of proposed Feedback-Network (FB-Net integrated with m-PCNN based feature fusion in Fig.1) to  CAM [17] and Grad-CAM [18] models applied to I-Net [2] and our FB-Net. For an unbiased comparison among the models, we calculated the detection accuracies of models listed below if and only if all the models' classification predictions are true positive for the given test data (i.e. multi-spectral test data with 7x16x16 is predicted as mega-solar power plant).  

\textbf{Compared Models for Solar Power Plant Class Activation Mapping :} 
\begin{quote}
\begin{itemize}
	\item  I-Net feature-averaging [2]
	\item I-Net using CAM in [17]
	\item I-Net using Grad-CAM  [18]
	\item feature averaging on FB-Net w/o gap layer
	\item FB-Net using CAM in [17]
	\item FB-Net w/o gap layer using Grad-CAM in [18]
	\item  \textbf{Proposed FB-Net with mPCNN based CAM} (see Fig.1 test case)
	\item FB-Net w/o gap layer with mPCNN based Grad-CAM
\end{itemize}
\end{quote}

Among the 105 positive samples in test data ( including solar power plant regions), 86 of them are predicted as solar power plant class for all the eight models compared in above list; therefore, for detection evaluation, we used these 86 samples. Class activation map results used to evaluate detection performances can be seen in Fig.6. For better visualization and seeing effect of m-PCNN fusion, feature representations are up-sampled with a scale factor of 16 using nearest neighbor assignment so that the detection results in Fig.6 are processed in 256x256 for all approaches including m-PCNN with linking size 15 for W (Eq.1).

For evaluation, we use Area Under Curve (AUC) metric [27, 28]. AUC is the area under the Receiver Operating Characteristic (ROC) curve, and ROC is curve of true positive rate (TPR) with respect to false positive rate (FPR) of detection (e.g. class activation maps) by changing the threshold value to compare with ground truth maps [27, 28]. AUC and ROC results of the compared models are given in Fig.7 and Fig.8. Proposed model in Fig.1, m-PCNN integration for class activation mapping to FB-Net, outperforms all other detection approaches applied to I-Net or FB-Net. The lowest detection performance is obtained from I-Net Cam model (AUC value is 0.6652). And, our proposed FB-Net with m-PCNN using the CAM weights (see Fig.1) has the best AUC performance (0.9571) for the given test data. In general, FB-Net based detection results outperform I-Net base approaches. Grad-CAM performances decreases because in a very few test samples, Grad-CAM computation for both I-Net and FB-Net yields zero gradient during backward process due to very small precision. In our system, GPU computation using Chainer Deep Learning library supports float32 bit precision. So, detection fails in these cases (see last column in Fig.6) when the gradient cannot be represented with float32 bit precision.

An interesting observation is that simple feature averaging performs better than common CAM [17] and Grad-CAM [18] on both I-Net and FB-Net. In a binary classification, activations of CNN features at the final convolutional layer seems to be very representative for the object of interest (e.g. solar power plants). Therefore, learned features result in quite reliable detection performance on test data, especially for our proposed FB-Net. These detection based performance evaluation of FB-Net demonstrate that using higher level features as a top-down feedback signal with the same bottom-up process in the same network (using shared weights as in Fig.1) can help richer representation of the class dependent activations. Therefore, it can yield better class activations around object region from the CNN features of the last convolutional layers.

\begin{figure}[h]
\begin{center}
\fbox{ \includegraphics[width=0.95\linewidth]{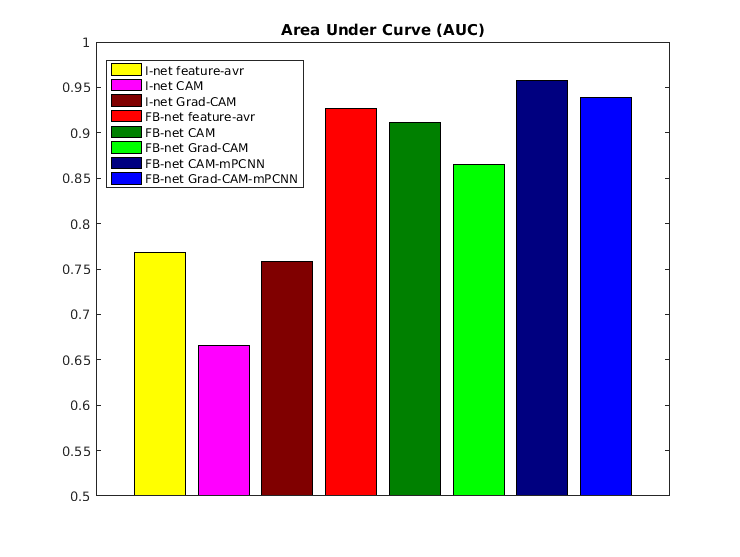}}
\end{center}
   \caption{Area Under Curve results for the compared models obtained from the ROC curve given in Fig. 8}
\label{fig:fig7}
\end{figure}

\begin{figure}[h]
\begin{center}
\fbox{ \includegraphics[width=0.95\linewidth]{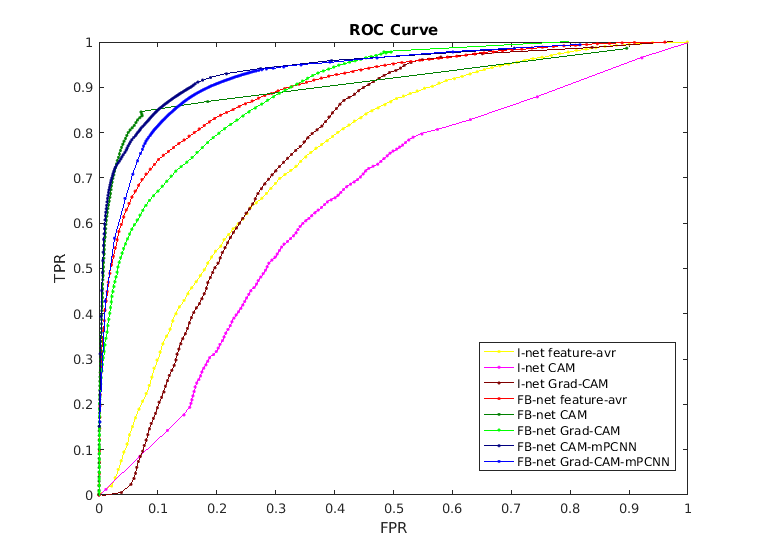}}
\end{center}
   \caption{Receiver Operating Characteristic curves obtained from the pixel wise comparison of the detection results with respect to ground truth for the compared modes}
\label{fig:fig8}
\end{figure}

\section{Conclusion}

In this work, we introduced a weakly supervised Feedback CNN to classify and detect solar power plants on multi-spectral data. For this, we took advantage of class activation mapping with m-PCNN on the proposed Feedback-CNN model, which can also be applied to larger scale multi-spectral scene (see sample results in Fig.9) without any resizing process. Experimental results demonstrated that CNN features extracted from the FB-Net are satisfactory for pixel-wise detection task (AUC metric) while providing good accuracy on classification based on IoU metric. 

\begin{figure}[h]
\begin{center}
\fbox{ \includegraphics[width=1.0\linewidth]{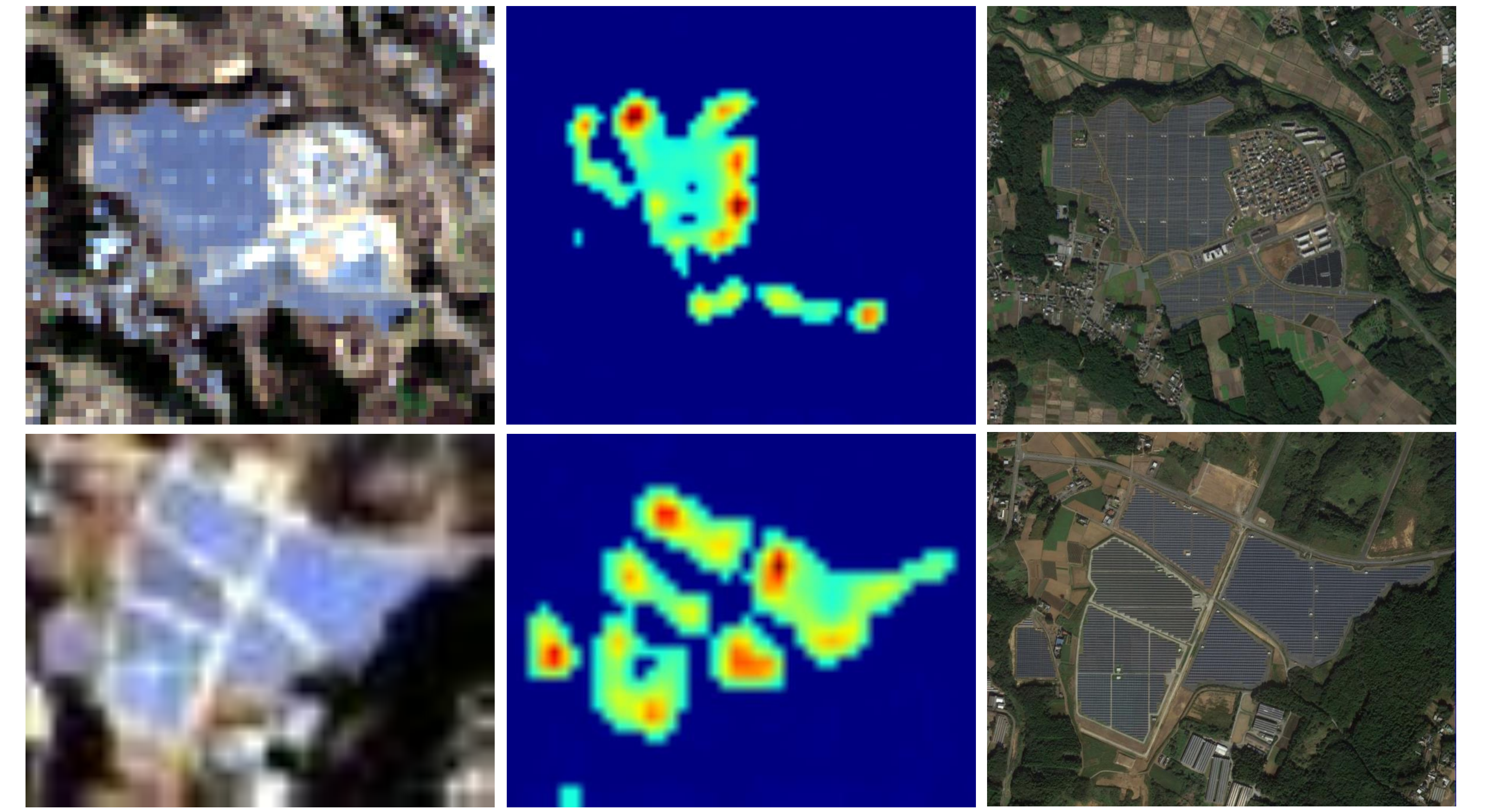}}
\end{center}
   \caption{(left) RGB visualization of input data (middle) CAM results obtained from proposed model in Fig.1 (right) Google Earth view of same area for reference}
\label{fig:fig9}
\end{figure}

As a future work, we are planning to try other baseline models to FB-Net and also try world-wide tests rather than limited region (Japan). Furthermore, we would like to extend the FB-Net implementation to other satellite imagery tasks such as change detection, damage detection, etc. In addition, we can investigate the effect of linking size in m-PCNN on detection accuracy. And, instead of setting parameters empirically, these parameters can be optimized with greedy search, genetic algorithm, or Bayesian optimization models.

\end{document}